%% file: main.tex
\documentclass[lettersize,journal]{IEEEtran}
\usepackage{amsmath,amsfonts}
\usepackage{algorithmic}
\usepackage{algorithm}
\usepackage{array}
\usepackage[caption=false,font=normalsize,labelfont=sf,textfont=sf]{subfig}
\usepackage{textcomp}
\usepackage{stfloats}
\usepackage{url}
\usepackage{verbatim}
\usepackage{graphicx}
\usepackage{cite}
\usepackage[colorlinks,linkcolor=red,anchorcolor=blue,citecolor=green]{hyperref}
\usepackage{color}
\usepackage{amsopn}
\usepackage{amssymb}
\usepackage{tabularx}
\usepackage{multirow}
\usepackage{booktabs}
\usepackage{bm}
\usepackage{xspace}
\usepackage{rotating}
\usepackage{epsfig}
\usepackage{balance}
\usepackage{makecell}
\usepackage{diagbox}
\usepackage{lineno,hyperref}

\begin{document}

\title{How to Understand Named Entities: \\Using Common Sense for News Captioning}

\author{Ning~Xu, Yanhui~Wang, Tingting~Zhang, Hongshuo~Tian, Mohan Kankanhalli~\IEEEmembership{Fellow,~IEEE}, and  An-An~Liu$^{\ast}$
\thanks{N.~Xu, Y.~Wang, T.~Zhang, H.~Tian, and A.-A.~Liu are with the School of Electrical and Information Engineering, Tianjin University, Tianjin 300072, China.
	M.~Kankanhalli is with the School of Computing, National University of Singapore, Singapore.
	(Corresponding author: A.-A.~Liu, E-mail: anan0422@gmail.com)}
\thanks{Manuscript received XXX, 202X; revised XXX, 202X.}}

\markboth{IEEE Transactions on Circuits and Systems for Video Technology,~Vol.~XX, No.~XX, December~2023}%
{Shell \MakeLowercase{\textit{et al.}}: A Sample Article Using IEEEtran.cls for IEEE Journals}

\IEEEpubid{0000--0000/00\$00.00~\copyright~2021 IEEE}

\maketitle

\input{sec_abstract}


\input{sec_introduction}
\input{sec_related_work}
\input{sec_approach}
\input{sec_experiment}
\input{sec_conclusion}
\input{sec_acknowledgement}

\bibliographystyle{IEEEtran} 
\bibliography{egbib}

\vfill

\end{document}

%% file: sec_abstract.tex
\begin{abstract}
News captioning aims to describe an image with its news article body as input.
It greatly relies on a set of detected named entities, including real-world people, organizations, and places.
{This paper exploits commonsense knowledge to understand named entities for news captioning.
By ``understand'', we mean correlating the news content with common sense in the wild, which helps an agent to 1) distinguish semantically similar named entities and 2) describe named entities using words outside of training corpora.}  
{Our approach consists of three modules: 
(a) \textit{Filter Module} aims to clarify the common sense concerning a named entity from two aspects: \textit{what does it mean?} and \textit{what is it related to?}, which divide the common sense into \textit{explanatory knowledge} and \textit{relevant knowledge}, respectively.
(b) \textit{Distinguish Module} aggregates \textit{explanatory knowledge} from \textit{node-degree}, \textit{dependency}, and \textit{distinguish} three aspects to distinguish semantically similar named entities.
(c) \textit{Enrich Module} attaches \textit{relevant knowledge} to named entities to enrich the entity description by commonsense information (e.g., identity and social position).
Finally, the probability distributions from both modules are integrated to generate the news captions.
Extensive experiments on two challenging datasets (i.e., GoodNews and NYTimes) demonstrate the superiority of our method.
Ablation studies and visualization further validate its effectiveness in understanding named entities.}
\end{abstract}

%% file: sec_introduction.tex
\section{Introduction}\label{sec:introduction}
\IEEEPARstart{D}{ifferent} from the traditional image captioning task \cite{VinyalsTBE15, XianLTM22, BenPLYHWM22, 9733170} that describes an image using general objects and their relationships, the news image captioning task aims to offer a plausible interpretation of images with specific people, organizations, and places (i.e., named entities) \cite{FengL13, BitenGRK19, TranMX20}.
Fig. \ref{fig:news_caption} provides an example to show the difference between both tasks.
Given an image with two persons talking with each other, a traditional image captioner may output the description ``Two men are talking with each other'', while a news image captioner can grab the factual knowledge behind the pixels to produce a more expressive description ``Bill Gates, chairman of Microsoft, is talking with Nicolas Sarkozy, the president of France''.
The news captioning task focuses more on how to integrate named entities for understanding and interpreting scenes \cite{BitenGRK19}.
Besides, this technique has enormous potential to assist readers in navigating news corpus with applications to news retrieval \cite{moradi2019approach, Chen17, BrownFJJY95} and personalized recommendations \cite{LeeLL02, LiCLS10, PatroBGGC20}.

\IEEEpubidadjcol
There have been some attempts to select named entities for describing news images.
For examples,
template-based methods \cite{BitenGRK19,HuCJ20} first produce template captions with placeholders, which are filled with named entities selected from the news articles.
Currently, several end-to-end solutions have also been proposed to capture named entities,
such as the sentence-level attention mechanism \cite{BitenGRK19}, the word-level progressive concentration \cite{HuCJ20}, and the byte-pair encoding \cite{TranMX20}.
However, this task still presents two key challenges for named entities understanding:

\input{./supplement/fig_news_caption}
\input{./supplement/fig_motivation}
\begin{itemize}
	\item \textit{How to distinguish semantically similar named entities in news articles.}
	As shown in Fig. \ref{fig:news_caption}, we have a part of text like this: ``Mr. Gates and Mr. Sarkozy would like to use the money to finance development in the world's poorest nations''.
	Existing captioning methods can extract the named entities ``Mr. Gates'' and ``Mr. Sarkozy'', but they cannot explicitly distinguish these two words and decide which word can be presented in the caption.
	This is a difficult challenge due to
	1) Both words belong to the same category, i.e., the famous people, leading to difficulty in separating from each other.
	2) Just depending on an image, a captioning model can not differentiate them.  
	3) These named entities are in the same sentence, which means they share a similar textual context. It further increases the difficulty to distinguish them.
	\item \textit{How to use the words outside of news articles to describe the named entities.}
	A news image covers several name entities, which need a large number of relevant and detailed semantics to interpret them, e.g., using ``the president of France'' to describe ``Mr. Sarkozy'', and using ``Bill Gates'' to enrich ``Mr. Gates''. Nevertheless, news articles may not provide these rich semantics, which limits models' ability to describe specific objects.
We calculate the proportion of words in a news image caption that also appear in the corresponding news article.
As shown in Tab. \ref{tab:recall},
we count the number of words that are shared by news captions and news articles, then calculate its ratio to the total number of words in news captions.
We find that only about 49\% and 69\% words can be contained in articles of GoodNews and NYTimes datasets, respectively.
It demonstrates that only depending on news articles cannot provide sufficient semantics to describe images.
Hence, we argue that a well-designed captioning model should be able to incorporate external words for image understanding. By ``external words'', we mean the words that are not presented in news articles but necessary for describing images, which can improve the accuracy and reliability of generated news captions.
\end{itemize}

\input{./supplement/tab_named_entity_recall}
To address above issues, it is reasonable to use common sense to understand named entities for news captioning.
Common sense is the widely held knowledge that can provide facts to help us explain the unknown named entities in the wild.
Therefore, a news captioning model could be attached to the commonsense knowledge base to accurately understand the named entities.
For examples, if a captioning model knows the fact ``Mr. Gates is a businessman'' and ``Mr. Sarkozy is a politician'', it may reasonably distinguish ``Mr. Gates'' and ``Mr. Sarkozy'' and contextually decide which person can be described in the caption.
If a captioning model learns that ``Nicolas Sarkozy served as President of France from 2007 until 2012'', it would present a better description ``Mr. Sarkozy, the president of France'', even though the news article does not provide such information.

{
In this paper, we propose a series of communicative commonsense-aware modules, 
which can use common sense beyond the given images and articles to understand named entities for news captioning task.
As shown in Fig. \ref{fig:motivation}, existing methods \cite{TranMX20, abs-2107-11970, LiuWWO21} use deep neural networks to encode images and articles, and then directly decode them for caption generation.
Comparatively, we additionally utilize common sense and divide it into two categories, i.e., \emph{explanatory knowledge} and \emph{relevant knowledge}.
The former examines a named entity by answering ``what does it mean?'', which provides a concept knowledge base to distinguish similar entities.
The latter analyzes a named entity by answering ``what is it related to?'', which provides more relevant semantics to completely describe an entity (e.g., identity and social position).
We try to decompose the traditional black-box deep neural networks into dynamic modules, which can teach machines how to ``use'' common sense in a flexible manner, regardless of how to ``memorize'' them.
}

{
As shown in Fig. \ref{fig:framework}, the pipeline of the proposed method consists of three novel communicative modules:
1) \textbf{Filter Module.} We first extract the set of named entities from the given article. Then, each entity is queried in ConceptNet to obtain its commonsense sub-graph, which is further divided into the \textit{explanatory} and \textit{relevant} ones for the distinguish and enrich modules, respectively.
2) \textbf{Distinguish Module.}
To distinguish the semantically similar named entities, we propose to enhance the entity representation by aggregating \textit{explanatory} sub-graph (i.e., knowledge) from three aspects, including \textit{node-degree}, \textit{dependency}, and \textit{distinguish} embeddings.
The multi-head attention computes the different weights in a probability distribution to decide which entity should be used for the sentence generation.
3) \textbf{Enrich Module.}
To describe a named entity more completely,
we propose the commonsense-entity interaction mechanism to associate each concept in the \textit{relevant} sub-graph with its entity, which enables a decoder to attach the reliable concepts for enriching the entity description.
Finally, we integrate the probability distributions from both the distinguish and the enrich modules to generate news captions.
}

We conduct extensive experiments on two popular news datasets, i.e., GoodNews \cite{BitenGRK19} and NYTimes \cite{TranMX20}, where our method achieves competitive performance against the state-of-the-art works. Especially, under the metrics of rare proper nouns, our method can obtain the significant performance improvements, which confirm the ability of understanding named entities. Qualitative results show that our method is highly explainable and explicit.

In summary, we make the following contributions:
\begin{itemize}
\item We are the first to utilize external common sense for named entity understanding in the news captioning task. Different from existing works that generate named entities only depending on news content, we define the key challenge of news captioning as ``how to use common sense that is beyond the given news content to infer named entities''.
\item We design three communicative modules to divide and integrate the commonsense knowledge, which present a series of explainable intermediate steps to distinguish semantically similar named entities and describe a named entity completely.
\item Extensive experiments on two popular news datasets show the superiority of our method. Ablation studies and qualitative analysis further validate the effectiveness.
\end{itemize}

%% file: supplement/fig_news_caption.tex
\begin{figure}
	\centering{\includegraphics[width=1.0\columnwidth]{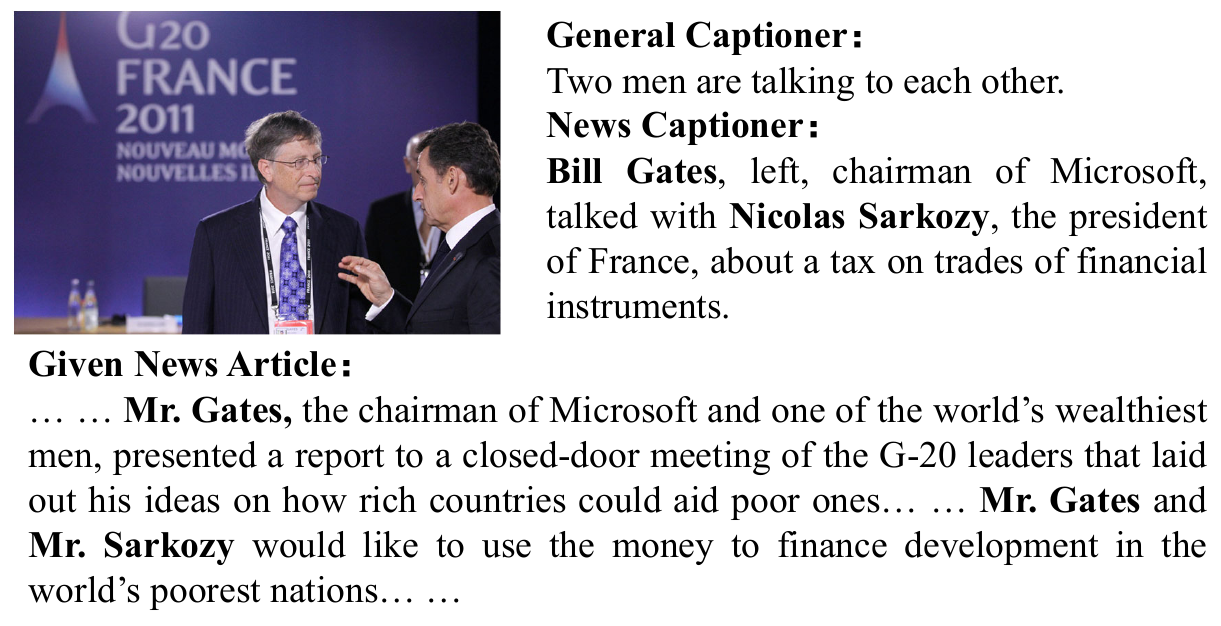}}
	\vspace{-0.6cm}
	\caption
	{Comparison of the general image caption and the news image caption.
The latter can produce more expressive descriptions with specific people, organizations, and places (i.e., named entities).}
	\vspace{-0.3cm}
	\label{fig:news_caption}
\end{figure}

%% file: supplement/fig_motivation.tex
\begin{figure*}
	\centering{\includegraphics[width=1.8\columnwidth]{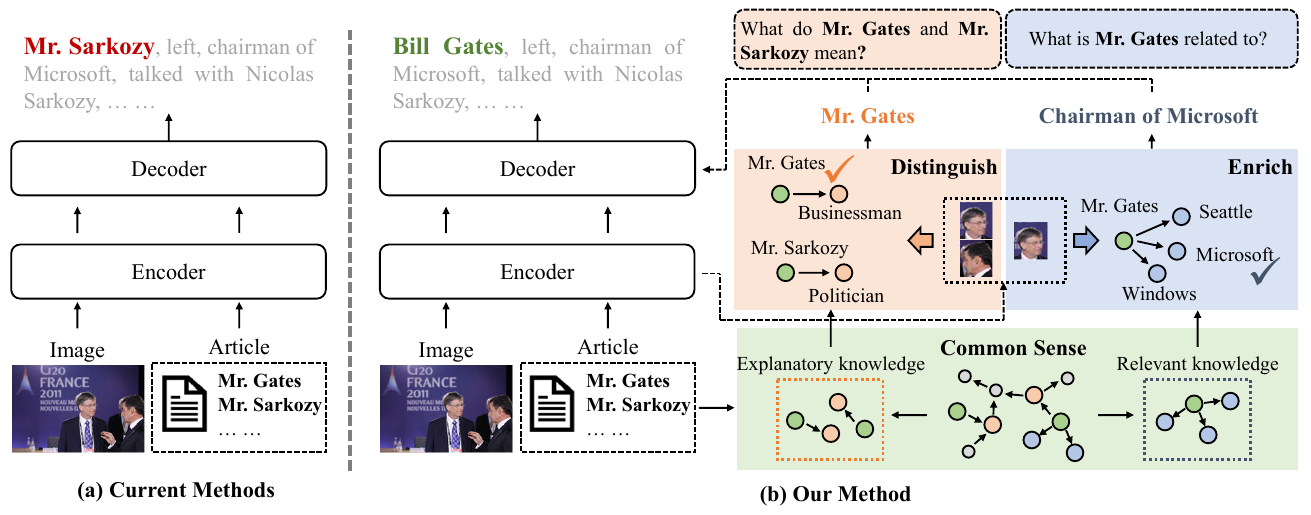}}
	\vspace{-0.3cm}
	\caption
	{
	Comparison of current methods and our method.
	Current methods generate news captions only relying on images and news articles.
	By comparison, we additionally use the external common sense and divide it into \emph{explanatory} and \emph{relevant} knowledge.
	The former is used to distinguish semantically similar named entities, while the latter is to provide more relevant semantics for the named entity description. 
	}
	\vspace{-0.3cm}
	\label{fig:motivation}
\end{figure*}

%% file: supplement/tab_named_entity_recall.tex
\begin{table}[!t]
	\centering
	\caption{Percentage of news captions' words that also appear in news articles on GoodNews and NYTimes datasets.
	The ratio indicates that only depending on news articles can not provide sufficient semantics to describe images.}
	\vspace{-0.1cm}
	\begin{tabular}{l|cc}
		\toprule
		 & GoodNews & NYTimes \\
		\midrule
		Train & 49.03\% & 69.87\% \\
		Validation & 49.03\% & 69.02\% \\
		Test & 49.18\% & 70.74\% \\
		\bottomrule
	\end{tabular}
	\vspace{-0.2cm}
	\label{tab:recall}
\end{table}

%% file: sec_related_work.tex
\section{Related Work}
\input{./supplement/tab_summarization}
\subsection{Image captioning}
Current captioning models \cite{XuBKCCSZB15, LuXPS17, RennieMMRG17, SongGGLHS19} mainly follow the encoder-decoder
framework, where a CNN-based encoder extracts visual representations from images and an RNN-based decoder utilizes the visual representations to generate the captions.

\noindent\textbf{Attention-based methods.}
{
To obtain rich and discriminative representations of images, recent works \cite{00010BT0GZ18, YouJWFL16, ChenZXNSLC17, PanYLM20} have introduced various attention mechanisms.
For examples,
visual attention mechanism \cite{00010BT0GZ18, LuXPS17, ChenZXNSLC17} attempted to focus on salient visual features of the images to generate more accurate words at each decoding step.
Moreover, semantic attention mechanism \cite{YouJWFL16, GaoLZSWS22, WangXLLZ22} proposed to selectively attend to semantic concepts that are  most relevant to the language context, which further improves the quality of generated sentences.
Based on previous works, Anderson \textit{et al}. \cite{00010BT0GZ18} combined the bottom-up visual attention with the top-down semantic attention to emphasize object-level regions for caption generation.
To avoid the negative effect of imposing attention on non-visual words such as ``the'' and ``a'', Gao \textit{et al}. \cite{GaoLSS20} propose an adaptive attention method, which automatically decides whether to depend on the visual information or the language context information for caption generation.
}

\noindent\textbf{Transformer-based methods.}
Sparked by the impressive breakthroughs \cite{VaswaniSPUJGKP17, DevlinCLT19, abs-1907-11692} of the Transformer architecture in the natural language processing field, many recent works \cite{YuLYH20, HuangWCW19, GuoLZYLL20, PanYLM20, XuMYTWJ22} proposed diverse Transformer-based models for the task of image captioning.
The core spirit of the Transformer-based captioning models is to capture the global dependencies between the visual features using the self-attention mechanism \cite{VaswaniSPUJGKP17}.
Particularly, Huang \textit{et al}. \cite{HuangWCW19} proposed to obtain visual representations of higher quality by filtering out irrelevant attended results via attention gates.
Guo \textit{et al}. \cite{GuoLZYLL20} extended the self-attention mechanism to explicitly model the relative geometry relations between the objects in the image.

Although having achieved superior performance \cite{DengDTW20, ZhangSLJZWHJ21} on the captioning benchmarks like MSCOCO \cite{LinMBHPRDZ14} and Conceptual Captions \cite{SoricutDSG18}, the general captioning systems can only produce common sentences that focus on the general image content, which cannot provide fine-grained information such as named entities for downstream applications \cite{NieQM00B22, YanHLYLMCG22, CaoAZW22}.

\subsection{News captioning}
The task of news captioning aims to describe the news images with named entities relying on the context information in the news articles.
Limited by the ability to represent multimodal information, early works focused on how to fully integrate different information sources  i.e., news images and news articles.
For examples,
Feng \textit{et al}. \cite{FengL13} introduced a knowledge-lean framework which can learn to select content and generate descriptions from weakly labeled data in an unsupervised fashion \cite{peng2023grlc, MoCLPSYZ23}.
Tariq \textit{et al}. \cite{TariqF17} proposed to use the probability space as the common representation space which can combine the information from different data spaces.
Ramisa \textit{et al}. \cite{RamisaYMM18} represented the context information using word2vec representations of the news articles and CNN visual features of news images.
However, previous methods suffered from generating named entities that were not seen during training.
Therefore, recent works focus more on how to generate correct named entities in description sentences.

\noindent\textbf{Template-based methods.}
To leverage the named entities appearing in the news articles, Template-based methods \cite{BitenGRK19, HuCJ20} first produce template captions with placeholders for named entities, which are filled with named entities selected from news articles.
For examples,
Biten \textit{et al}. \cite{BitenGRK19} adopted a two-stage manner, which finishes template generation first and chooses named entities with attention values to fill the template in the second stage.
In order to select the named entities more accurately, Hu \textit{et al}. \cite{HuCJ20} proposed to selectively focus on the relevant sentences in the news article and jointly optimize the template caption generation and the named entity selection in an end-to-end manner.
However, template-based methods limit the linguistically richness of generated captions and are vulnerable to errors in template generation.

\noindent\textbf{Joint learning methods.}
To overcome the above limitation, Joint learning methods  \cite{TranMX20,abs-2107-11970,LiuWWO21} directly generate the named entities and linguistically rich sentences without the selection process and template captions.
Specially,
Alasdair \textit{et al}. \cite{TranMX20} exploited a Transformer decoder equipped with byte-pair-encoding, which enables to generate unseen or rare named entities using a sequence of subword units.
Zhao \textit{et al}. \cite{abs-2107-11970} facilitated the caption generation by constructing a multi-modal knowledge graph that matches textual entities in news articles and visual objects of images.
However, existing methods still suffer from understanding named entities due to the lack of necessary commonsense knowledge.


\subsection{Exploiting Common Sense}
{
Exploiting external common sense to supplement missing information in context has been a crucial approach in vision and language tasks.
Yang \textit{et al}. \cite{YangM17} is among the first to retrieve related entities from knowledge bases and merge their embeddings into the language representations, which improves the performance on the task of entity/event extraction.
Following the line of works, many works \cite{FrankM18, AnnervazCD18} have attempted to incorporate the embeddings of relevant knowledge triplets at the word level in natural language understanding tasks.
ConceptNet presents commonsense knowledge in the form of structural graph, which serves as a reliable commonsense source for recent works.
For example, Lin \textit{et al}. \cite{LinCCR19} emphasizes the infusion of external knowledge in an explicit graph structure while preserving the relationship reasoning ability of the graph.
Yasunaga \textit{et al}. \cite{YasunagaRBLL21} connects the context in question-answer pair and knowledge graph to form a joint graph, then mutually updates their representations through graph neural networks.

However, previous methods always ignore the semantic differences among diverse commonsense concepts and indiscriminately merge them with the language context, which can introduce unnecessary noise and constrain the model performance.
}

\subsection{Summarization}
Table \ref{tab:summary} presents a comprehensive comparison between current news captioning models and our method.
Different from the existing models, which employ the template-based or joint learning approaches,
we utilize the external common sense that is beyond the given news content (i.e., article and image) to infer named entities for sentence generation.
Our method aims to explicitly distinguish semantically similar named entities and describe a named entity completely, which can provide a series of explainable intermediate steps to improve the named entity understanding.


%% file: supplement/tab_summarization.tex
\begin{table*}[!t]
	\centering
	\caption{Comparison of news captioning methods.}
	\vspace{-0.1cm}
	\begin{tabular}{l|ccc|c}
		\toprule
		\textbf{Ref.} & \textbf{Template-based} & \textbf{Joint Learning} & \textbf{Common Sense} & \textbf{Method}                                    \\
		\midrule
		\makecell[l]{\cite{BitenGRK19}, \cite{XuBKCCSZB15}, \cite{00010BT0GZ18},\\ \cite{RamisaYMM18}, \cite{HuCJ20} + CtxIns} & \checkmark  	& 				& & \makecell[l]{Fill the template captions by selecting named entities \\from news article.} \\
		\midrule
		\cite{TranMX20}, \cite{abs-2107-11970},\cite{LiuWWO21}          & 				& \checkmark 	& & \makecell[l]{Extract and integrate high-level attributes from news content\\ to produce sentences in an end-to-end fashion.} \\
		\midrule
		\textbf{Ours}        & 				& \checkmark 	& \checkmark & \makecell[l]{Use common sense that is beyond the given news content \\to infer named entities for sentence generation.} \\
		\bottomrule
	\end{tabular}
	\vspace{-0.2cm}
	\label{tab:summary}
\end{table*}

%% file: sec_approach.tex
\input{./supplement/fig_framework}
\section{Approach}\label{sec:approach}
\subsection{Problem Formulation}
{
In this section, we formally introduce the news captioning task and the idea of using common sense to understand named entities.
Given a news image $\boldsymbol{I}$ and its associated article $\boldsymbol{A}$, the task of news captioning aims to generate a sentence $\boldsymbol{S} = \{\boldsymbol{w}_1, \boldsymbol{w}_2, ... , \boldsymbol{w}_T\}$ with rich named entities, which can be defined by:
\begin{equation}{\label{eq:problem_formulation}}
\log p(\boldsymbol{S} \mid \boldsymbol{I}, \boldsymbol{A})=\sum_{t=1}^{T} \log p\left(\boldsymbol{w}_{t} \mid \boldsymbol{I}, \boldsymbol{A}, \boldsymbol{w}_{0:t-1}\right)
\end{equation}
where $p(\cdot)$ is the probability distribution over the set of possible words; $\boldsymbol{w}_{0:t-1} = \{\boldsymbol{w}_1, \boldsymbol{w}_2, ... , \boldsymbol{w}_{t-1}\}$ denotes the words generated before the time step $t$; $T$ is the sentence length.

To distinguish semantically similar named entities, and meanwhile describe named entities using knowledge outside of training corpora, we use the common sense to improve the named entity generation for news captioning.
Formally, the probability function in Eq. \ref{eq:problem_formulation} is decomposed as follows:
\begin{equation}
\begin{split}
p\left(\boldsymbol{w}_{t} \mid \boldsymbol{I}, \boldsymbol{A}, \boldsymbol{w}_{0:t-1}\right) = (1-\alpha) * p_{g}\left(\boldsymbol{w}_{t} \mid \boldsymbol{I}, \boldsymbol{A}, \boldsymbol{w}_{0:t-1}\right) \\ + \alpha * p_{n}\left(\boldsymbol{e}_{t} \mid \boldsymbol{I}, \boldsymbol{A}, \boldsymbol{C}, \boldsymbol{w}_{0:t-1}\right)
\end{split}
\end{equation}
where $p_{g}$ denotes the probability distributions of general words such as ``man'', which can describe the general visual signals of an image, but it is agnostic to named entities.
Comparatively, we propose the named entity distribution $p_{n}$, which leverages the commonsense knowledge $C$ to increase the probabilities of named entities, such as ``Bill Gates'';
$e_t$ denotes the named entity generated at the time step $t$;
$\alpha$ is a learned parameter that balances the probability between a general word and a named entity.}

{
Technically, we propose three communicative commonsense-aware modules, which present a series of explainable intermediate steps for named entity understanding:
1) \textbf{Acquiring commonsense knowledge:} The filter module $\operatorname{O}_{f}$ uses the named entities to retrieve common sense, which is further refined to satisfy the needs of the following modules;
2) \textbf{Distinguishing similar named entities: } The distinguish module $\operatorname{O}_{d}$ captures discriminative cues in commonsense
to serve accurate probability distribution for semantically similar named entities;
3) \textbf{Enriching named entities description:} The enrich module $\operatorname{O}_{e}$ extends the probability distribution depending on common sense to describe named entities more completely.}
Accordingly, the named entity distribution $p_{n}$ is re-written as:
\begin{equation}
\begin{split}
p_{n}\left(\boldsymbol{e}_{t} \mid \boldsymbol{I}, \boldsymbol{A}, \boldsymbol{C}, \boldsymbol{w}_{0:t-1}\right) =
\beta * \operatorname{O}_d\left(\boldsymbol{I}, \boldsymbol{A}, \operatorname{O}_{f}\left(\boldsymbol{A}, \boldsymbol{C}\right), \boldsymbol{w}_{0:t-1}\right) \\+ (1-\beta) * \operatorname{O}_e\left(\boldsymbol{I}, \boldsymbol{A}, \operatorname{O}_{f}\left(\boldsymbol{A}, \boldsymbol{C}\right), \boldsymbol{w}_{0:t-1}\right)
\end{split}
\end{equation}
where $\beta$ is a learned parameter that balances the probability distributions generated by the distinguish and enrich modules.

\subsection{Pre-processing}
We first use the pre-trained ResNet-152 \cite{HeZRS16} to encode the given image $\boldsymbol{I}$.
The block before the last pooling layer divides the image into $N^p$ patches, and
outputs the patch set $\boldsymbol{X}^{(p)} = \{\boldsymbol{x}^{(p)}_1, ..., \boldsymbol{x}^{(p)}_{N^p}\}$, where $\boldsymbol{x}^{(p)}_{n^p}$ is the feature of the $n^p$-th image patch; $N^p$ is the number of image patches.
Similar to \cite{TranMX20}, we use MTCNN \cite{ZhangZLQ16} and FaceNet \cite{SchroffKP15} to extract people's face features of an image to obtain the face set $\boldsymbol{X}^{(f)} = \{\boldsymbol{x}^{(f)}_1, ..., \boldsymbol{x}^{(f)}_{N^f}\}$, where $N^f$ is the number of faces.
We use YOLOv3 \cite{abs-1804-02767} to extract object features and output the object set $\boldsymbol{X}^{(o)} = \{\boldsymbol{x}^{(o)}_1, ..., \boldsymbol{x}^{(o)}_{N^o}\}$, where $N^o$ is the number of objects.
These three types of image features are collected to form the image representation $\boldsymbol{X}^I = \{ \boldsymbol{X}^{(p)}, \boldsymbol{X}^{(f)}, \boldsymbol{X}^{(o)} \}$.
Meanwhile, we use the pre-trained RoBERTa \cite{abs-1907-11692} to represent the associated news article $\boldsymbol{A}$. We denote the given article has $T$ words.
The RoBERTa first encodes $\boldsymbol{A}$ into a set of sequences $\{w_l^{(t)}:l\in\{1,2,\ldots,L\},t\in\{1,2,\ldots,T\}\}$, where $L$ is the length of each sequence.
Similar to \cite{TranMX20, TenneyDP19}, we take a weighted sum to calculate the embedding of each word $\boldsymbol{x}^A_t = \Sigma^L_{l=0} \alpha_l w_l^{(t)} $, where $\alpha_l$ refers to the learned weight.
It produces a set of token vectors $\boldsymbol{X}^A = \{\boldsymbol{x}^A_1, \boldsymbol{x}^A_2, ..., \boldsymbol{x}^A_T\}$ as the article representation.

\input{./supplement/tab_relations}
\subsection{Dataset Selection}
Many commonsense datasets (e.g., Cyc, Microsoft Concept Graph, and ConceptNet) provide a rich variety of knowledge for named entities, such as ``people's name'' for ``Lionel Messi''.
However, a named entity essentially needs some factual knowledge to describe its particular meanings.
For examples, both of ``Lionel Messi'' and ``Lebron James'' belong to the knowledge ``people's name'', but ``people's name'' cannot help us understand their differences.
In contrast, factual knowledge like ``Lionel Messi is a football player'' and ``Lebron James is a basketball player'' counts more when comparing these two similar named entities.
Hence, it is critical to select a dataset that contains sufficient factual knowledge as our common sense bases.
To this end, we review several widely used datasets and select ConceptNet \cite{SpeerCH17} as the external knowledge base.
ConceptNet collects factual knowledge from a variety of resources, including expert-created resources (e.g., WordNet and JMDict), crowd-sourced resources (e.g., Wiktionary and Open Mind Common Sense), and games with a purpose (e.g., Verbosity and nadya.jp). Statistically, it provides 21 million facts involving 8 million concepts in total, which is conducive to distinguishing and describing named entities.

\subsection{Commonsense-Aware Modules}


The proposed method consists of the following three communicative modules.

\subsubsection{Filter Module}
It aims to query each named entity of ConceptNet and obtain the commonsense sub-graph, which is further divided into the \textit{explanatory} and \textit{relevant} sub-graphs for the following distinguish and enrich modules, respectively.

\input{supplement/fig_filter_module.tex}
\noindent\textbf{Common Sense Extraction.}
Before diving into the extraction of commonsense knowledge, we would like to briefly introduce the used ConceptNet \cite{SpeerCH17} dataset.
ConceptNet is the large-scale structured knowledge base that describes concepts and their commonsense relations in the form of $<$subject-relation-object$>$ triplets, such as $<$Bill Gates-IsA-Businessman$>$ and $<$Microsoft-CreatedBy-Bill Gates$>$.
One can start using ConceptNet by querying a concept which returns a sub-graph that represents the concept-relevant knowledge.

{
Existing methods \cite{LinCCR19,LvGXTDGSJCH20,ShwartzWBBC20} always get this sub-graph by retrieving paths in ConceptNet to find a sequence of triplets.
For examples, given a named entity (i.e., query) ``Bill Gates'', the returned sub-graph not only link to ``Microsoft'' via the fact ``Bill Gates is the founder of Microsoft'', but also associates with ``Apple'' and ``Steve Jobs'' due to that ``Microsoft is similar to Apple'' and ``Apple is created by Steve Jobs''.
Then, the sequential triplets including ``Bill Gates $\rightarrow$ Microsoft $\rightarrow$ Apple $\rightarrow$ Steve Jobs'' is considered as the multi-hop knowledge.
Due to the tremendous size of ConceptNet, it is easy to introduce semantic noise and increase the computational burden in the multi-hop way.
Meanwhile, we observe that only using the one-hop retrieval results (e.g., ``Bill Gates $\rightarrow$ Microsoft'') could provide sufficient knowledge facts to explain most of named entities.
Therefore, we adopt a straightforward yet effective method to get the sub-graph of ConceptNet for an entity.
We first use the tool of SpaCy\footnote{https://spacy.io/} to extract the set of named entities from the news article, which is denoted by $\{e_k\}^K_{k=1}$ and $K$ is the number of entities.
As shown in Fig. \ref{fig:filter}, we retrieve these entities by searching for paths in the one-hop range.
Particularly, we consider an entity $e_k$ may participate in many triplets.
If an entity is the head of a triplet, we collect the tail concepts of all such triplets to build the set $\tilde{C}^h_k$.
Similarly, if an entity is the tail of a triplet, we obtain the head concepts of all such triplets to build the set $\tilde{C}^t_k$ as follows:
\begin{equation}
\tilde{C}^h_k = \{tail_i:(\boldsymbol{e}_k, relation, tail_i) \in ConceptNet \}
\end{equation}
\begin{equation}
\tilde{C}^t_k = \{head_j:(head_j, relation, \boldsymbol{e}_k) \in ConceptNet \}
\end{equation}
The sub-graph retrieved from ConceptNet for an entity $e_k$ is build as $\tilde{C}_k = \{\tilde{C}^h_k \cup \tilde{C}^t_k\}$.

Further, we use the method of TransE \cite{BordesUGWY13} to prune irrelevant paths in the sub-graph.
We decompose it into a set of triplets, which confidence is directly measured by the scoring function of TransE.
A threshold of triplet classification is empirically set for pruning and finally we get the updated sub-graph $C_k$.

\noindent\textbf{Common Sense Division.}
We seek the entity-aware insights on the sub-graph $C_k$.
To capture the informative concepts to distinguish and describe named entities, we categorize the relations of  commonsense knowledge into two high-level types.
As shown in Table \ref{tab:relations}, the \textit{explanatory} relations can help searching for the concepts that reflect the \textit{inherent meaning} of a named entity.
For examples, the relation ``IsA'' connects the named entity ``Bill Gates'' with a concept ``Businessman'', ``Nicolas Sarkozy'' with a concept ``Politician''.
These concepts can provide the factual evidence to distinguish ``Bill Gates'' and ``Nicolas Sarkozy''.
In contrast, the \textit{relevant} relations aim to find the concepts that denote the \textit{extended meaning} of a named entity.
For examples, the relation ``CreatedBy'' points out that the named entity ``Bill Gates'' found the company ``Microsoft'', which introduces the rich semantic to produce the description ``the father of Microsoft, Bill Gates''.

Using our high-level relation types, the graph $C_k$ can be further divided into the \textit{explanatory} and \textit{relevant} common sense sub-graphs, denoted by $C^e_k$ and $C^r_k$, respectively.
Both sub-graphs provide the supporting facts to restrict that the named entity understanding can be drawn from a fixed and stable semantic space. 
}

\subsubsection{Distinguish Module}\label{sec:distinguish}
This module aims to depend on common sense to analyze semantically similar named entities (such as ``Bill Gates'' and ``Nicolas Sarkozy'') and decide which one should be used for the sentence generation.

From the previous module, we can obtain the \textit{explanatory} commonsense sub-graph $C^e_k$ for an entity. To effectively distinguish the similar entities, we propose to enhance the named entity representation by aggregating the commonsense sub-graph.
Firstly, as shown in Fig. \ref{fig:framework}, we pick out all of concepts in sub-graph $C^e_k$, where
the features of entity and concepts are extracted by RoBERTa \cite{abs-1907-11692}, denoted as $x_n^{(k)}$.
We build the entity-commonsense sequence $\boldsymbol{X}_k = \{\boldsymbol{x}_0^{(k)}, \boldsymbol{x}_1^{(k)}, \ldots, \boldsymbol{x}_n^{(k)}, \ldots, \boldsymbol{x}_N^{(k)}\}$, where $\boldsymbol{x}_0^{(k)}$ is the entity feature and $\boldsymbol{x}_{1,\ldots,N}^{(k)}$ are the concept features from commonsense.
We consider aggregating \textit{explanatory} sub-graph $C^e_k$ from the following three aspects:

\noindent \textbf{a) Node-Degree Embedding.}
In our commonsense graph, if a concept (i.e., node) links to the named entity via more edges than others, we say that this node may be a stronger signal for entity understanding.
The node degree is the number of edges that are incident to the node, which can measure how important a node is in the graph.
Nevertheless, such information is not directly reflected in the traditional attention calculation \cite{BahdanauCB14}.
We assign the real-valued embedding vector to encode each node degree as follows:
\begin{equation}
\tilde{\boldsymbol{x}}_n^{(k)} = \boldsymbol{x}_n^{(k)} + \boldsymbol{z}_{deg(n)}
\end{equation}
where $\boldsymbol{z}_{deg(n)}$ is a learnable vector according to the sum of indegree and outdegree of the $n$-th node; $\tilde{\boldsymbol{x}}_n^{(k)}$ is the updated vector after embedding.

\input{supplement/fig_distinguish_module.tex}
\noindent \textbf{b) Dependency Embedding.}
The semantic dependency between nodes in our commonsense graph is an important factor for named entity understanding.
For example, the entity ``Bill Gates'' is linked to ``Businessman'' and ``American'' with the high-score edge weights (the weights are provided by ConceptNet), which can explicitly specify the semantic dependency via statistical co-occurrences.
To make use of the dependency information, we re-arrange the entity-commonsense sequence $\boldsymbol{X}_k$ by ordering \textit{explanatory} concepts in descending order of edge weights.
Specifically, if a concept has a higher edge weight linked to the named entity than the other, then it should come before the other.
So, we obtain the new entity-commonsense sequence $\boldsymbol{\bar{X}}_k = \{\boldsymbol{\bar{x}}_0^{(k)}, \boldsymbol{\bar{x}}_1^{(k)}, \ldots, \boldsymbol{\bar{x}}_N^{(k)}\}$.
Referring to \cite{VaswaniSPUJGKP17}, we add ``positional encoding'' to each concept feature for memorizing the sequence order:
\begin{equation}
\begin{aligned}
z_{n, 2i} &=\sin \left(n / 10,000^{2i / d}\right) \\
z_{n, 2i+1} &=\cos \left(n / 10,000^{2i / d}\right)
\end{aligned}
\end{equation}
where $n$ is the concept position and $i \in [0, d/2]$ is the feature dimension. That is, each dimension of the positional encoding corresponds to a sinusoid.
In this way, we give each position an embedding to refine the concept feature:
\begin{equation}
\boldsymbol{\check{x}}_n^{(k)} = \boldsymbol{\bar{x}}_n^{(k)} + \boldsymbol{z}_{pos(n)}
\end{equation}
where $\boldsymbol{z}_{pos(n)}$ is a learnable vector for the positional embedding of the $n$-th node; $\boldsymbol{\check{x}}_n^{(k)}$ is the refined vector.

\noindent \textbf{c) Distinguish Embedding.}
When comparing two similar named entities, we often look for the decisive concepts in the \textit{explanatory} commonsense to distinguish them.
For example, to distinguish ``Mr. Gates'' and ``Mr. Sarkozy'', the ``businessman'' and ``politician'' are more effective and believable concepts to separate them, than the shared ``famous man''.
In other words, ``businessman'' and ``politician'' are significantly irrelevant, which can facilitate the distinguishing process.
Therefore, it is essential to embed the irrelevant information for the named entity representation.
Specifically, given two entities $e_k$ and $e_g$, we propose to calculate the semantic similarity scores between their \textit{explanatory} commonsense graphs $C^e_k$ and $C^e_g$.
Particularly, for each concept in $C^e_k$, we compute its similarity score with all concepts in $C^e_g$ by the function spaCy \cite{honnibal2017spacy}, where
$s_{n,m}$ denotes the score between the $n$-th concept in $C^e_k$ and the $m$-th concept in $C^e_g$.
Then, we sum up the similarity scores to obtain $s_n = \sum_{m} s_{n,m}$ that reflects the degree of similarity between the $n$-th concept in $C^e_k$ and the entity $e_g$.
Therefore, its reciprocal $\frac{1}{s_n}$ measures the degree of irrelevance,
which can be used to scale the feature of the $n$-th concept in $C^e_k$:
\begin{equation}
\boldsymbol{\hat{x}}^{(k)}_n = \boldsymbol{\check{x}}^{(k)}_n * \frac{1}{s_n}
\end{equation}
where $\boldsymbol{\hat{x}}^{(k)}_n$ is the scaled vector.

%

Till now, we obtain the refined entity-commonsense sequence $\boldsymbol{\hat{X}}_k= \{\boldsymbol{\hat{x}}_0^{(k)}, \boldsymbol{\hat{x}}_1^{(k)}, \ldots, \boldsymbol{\hat{x}}_n^{(k)}, \ldots, \boldsymbol{\hat{x}}_N^{(k)}\}$,
which are averaged to get the enhanced named entity representation
$\boldsymbol{\hat{x}_k} = \frac{1}{N+1} \sum_{n} \boldsymbol{\hat{x}}_n^{(k)}$.
Then, all of the named entities are collected to build the entity matrix
$\boldsymbol{X}^{E} = \{ \boldsymbol{\hat{x}_1}, \ldots, \boldsymbol{\hat{x}_K} \}$.

To decide which entity should be used for the sentence generation, we use the multi-head attention \cite{VaswaniSPUJGKP17} to assign the different weights in a probability distribution for similar entities:
\begin{equation}
\boldsymbol{\alpha}_{t}^{E}=\operatorname{MultiHead}\left(
\boldsymbol{W}^e_Q\boldsymbol{m}_{t}, \boldsymbol{W}^e_K\boldsymbol{X}^{E}, \boldsymbol{W}^e_V\boldsymbol{X}^{E}
\right)
\end{equation}
where $\boldsymbol{W}^e_Q$, $\boldsymbol{W}^e_K$, $\boldsymbol{W}^e_V$ are learnable matrices;
$\boldsymbol{\alpha}_{t}^{E}$ is the entity probability distribution at time step $t$, where the high scorer is more likely to appear in the output caption;
$\boldsymbol{m}_{t}$ is the context representation vector which is computed as follows:

\begin{equation}\label{eq:trans_layer}
\boldsymbol{m}_{t}^{l} =\operatorname{Layer}^{l}\left(\boldsymbol{m}_{t}^{l-1},  \boldsymbol{m}_{0:t-1}^{l-1}, \boldsymbol{X}^{I}, \boldsymbol{X}^{A}, \boldsymbol{X}^{E}, \boldsymbol{X}^{R}\right)
\end{equation}
where $\operatorname{Layer}^{l}$ is the $l$-th decoder layer that belongs to a Transformer module \cite{VaswaniSPUJGKP17};
$\boldsymbol{m}_{t}^{l}$ is the intermediate vector at time step $t$;
$\boldsymbol{m}_{t}^{l-1}$ is the output vector of previous layer at time step $t$;
$\boldsymbol{m}_{0:t-1}^{l-1}$ is the output vectors obtained in previous time steps;
$\boldsymbol{X}^{I}$ and $\boldsymbol{X}^{A}$ are the image and article representation, respectively;
$\boldsymbol{X}^{R}$ is the commonsense representation (will be explained in Enrich Module).
Specifically, each decoder layer consists of two steps:
\begin{equation}
\begin{aligned}
\tilde{\boldsymbol{m}}_{t}^{l} = &\operatorname{LayerNorm}(
\\ \boldsymbol{m}_{t}^{l-1} +
&\operatorname{MultiHead}(
\boldsymbol{\tilde{W}}_Q\boldsymbol{m}_{t}^{l-1}, \boldsymbol{\tilde{W}}_K\boldsymbol{m}_{0:t-1}^{l-1}, \boldsymbol{\tilde{W}}_V\boldsymbol{m}_{0:t-1}^{l-1}
))
\end{aligned}
\end{equation}

\begin{equation}
\begin{aligned}
\boldsymbol{m}_{t}^{l+1} = &\operatorname{LayerNorm}(
\\ \tilde{\boldsymbol{m}}_{t}^{l} +
&\operatorname{MultiHead}(
\boldsymbol{\bar{W}}_Q\tilde{\boldsymbol{m}}_{t}^{l},
\boldsymbol{\bar{W}}_K \boldsymbol{X}_{co}, \boldsymbol{\bar{W}}_V\boldsymbol{X}_{co}
))
\end{aligned}
\end{equation}
where $\boldsymbol{\tilde{W}}_Q$, $\boldsymbol{\bar{W}}_Q$, $\boldsymbol{\tilde{W}}_K$, $\boldsymbol{\bar{W}}_K$, $\boldsymbol{\tilde{W}}_V$, $\boldsymbol{\bar{W}}_V$ are learnable matrices;
$\boldsymbol{X}_{co} = [\boldsymbol{X}^{I};\boldsymbol{X}^{A}; \boldsymbol{X}^{E};\boldsymbol{X}^{R}]$ means that four types of information are concatenated, which is fed through a residual connection and layer normalization.
The final output $\boldsymbol{m}^{L}_{t}$ ($L$=24) is used as the context representation vector $\boldsymbol{m}_{t}$.

\subsubsection{Enrich Module}\label{sec:enrich}
This module aims to analyze common sense and provide more relevant semantics that are beyond the given article, to describe a named entity completely (such as identity and social position).
For example, the entity ``Bill Gates'' is associated with his corporation ``Microsoft'', which can contribute to generating more fine-grained descriptions such as ``Bill Gates, the chairman of Microsoft'', even if the news article does not cover such detailed information.

We can obtain the \textit{relevant} commonsense sub-graph $C^r_k$ for an entity by the Filter Module.
Different from the \textit{explanatory} commonsense, which is the inherent meaning for a named entity (e.g., ``Bill Gates'' is a ``businessman''), the \textit{relevant} commonsense can provide the extended meaning around an entity (e.g., ``Bill Gates'' created ``Microsoft''). 
As described above, the distinguish module aims to separate similar entities by aggregating \textit{explanatory} commonsense to enhance the entity representation.
On the contrary, in this module, we focus on the role of commonsense, which could be used to produce the sentence. Therefore,
we propose to aggregate the named entities to enhance the \textit{relevant} commonsense representation.

Specifically, we pick out all of concepts in sub-graph $C^r_k$, which features are extracted by RoBERTa \cite{abs-1907-11692}, denoted as $c_n^{(k)}$ that means the $n$-th related concept for the $k$-th entity.
Then, each concept feature is followed by the corresponding entity feature $\boldsymbol{e}_{k}$.
All of the concepts are stacked to form the commonsense-entity sequence:
\begin{equation}
\bar{C}=\left(
\left[\begin{array}{l}
c_{1}^{(1)} \\
e_{1}
\end{array}\right],
\left[\begin{array}{c}
c_{2}^{(1)} \\
e_{1}
\end{array}\right], \ldots,
\left[\begin{array}{l}
c_{1}^{(2)} \\
e_{2}
\end{array}\right],
\left[\begin{array}{l}
c_{2}^{(2)} \\
e_{2}
\end{array}\right],
\ldots\right)
\end{equation}
We use the multi-head attention to refine the commonsense representation as follows:
\begin{equation}
\boldsymbol{X}^{R} = \operatorname{MultiHead}\left(
\boldsymbol{\bar{W}}^c_Q \bar{C}, \boldsymbol{\bar{W}}^c_K \bar{C}, \boldsymbol{\bar{W}}^c_V \bar{C}
\right)
\end{equation}
where $\boldsymbol{\bar{W}}^c_Q$, $\boldsymbol{\bar{W}}^c_K$, $\boldsymbol{\bar{W}}^c_V$ are learnable matrices. In this way, each concept can be individually associated with its entity, and the commonsense-entity interaction enables a decoder to attach reliable concepts to describe entities.

Similar to the distinguish module, we use another multi-head attention as the decoder to integrate the context representation $\boldsymbol{m}_{t}$ to produce a probability distribution for concepts:
\begin{equation}
\boldsymbol{\alpha}_{t}^{R}=\operatorname{Multi-Head}\left( \boldsymbol{W}^c_Q \boldsymbol{m}_{t}, \boldsymbol{W}^c_K \boldsymbol{X}^{R}, \boldsymbol{W}^c_V \boldsymbol{X}^{R}\right)
\end{equation}
where $\boldsymbol{W}^c_Q$, $\boldsymbol{W}^c_K$, $\boldsymbol{W}^c_V$ are learnable matrices;
$\boldsymbol{\alpha}_{t}^{R}$ is the probability distribution of commonsense concepts at time step $t$, in which the high scorer is more likely to be used to enrich the description of named entities.

\subsection{Probability Distribution Integration}\label{sec:enrich}
We integrate the probability distributions from both the distinguish module and the enrich module, to generate news captions.
The former helps to distinguish similar entities while the latter provides sufficient and relevant semantics to describe an entity.
We first use the context representation vector $\boldsymbol{m}_{t}$ to produce
the baseline probability distribution of the $t$-th word in the vocabulary $p\left(w_{t}\right)$ via an adaptive softmax \cite{GraveJCGJ17}:

\begin{equation}
p\left(w_{t}\right)=\operatorname{AdaptiveSoftmax}\left(\boldsymbol{W}_p \boldsymbol{m}_{t}\right)
\end{equation}
where $\boldsymbol{W}_p$ is the learnable matrix.
Then, the entity distribution $\boldsymbol{\alpha}_{t}^{E}$ and the commonsense distribution $\boldsymbol{\alpha}_{t}^{R}$ are collaboratively used to adjust the current word, to progress towards the goal of using commonsense for named entity understanding:

\begin{equation}\label{eq:fwd}
p^*\left(w_t\right)= x \ast \boldsymbol{\alpha}_{t}^{E}+ y \ast \boldsymbol{\alpha}_{t}^{R} + (1-x-y) \ast p\left(w_t\right)
\end{equation}
where $p^*\left(w_t\right)$ is the final distribution to predict the current word.
$\boldsymbol{\alpha}_{t}^{R}$ is used as the attention distribution to tell the generator which commonsense concepts should be copied for the word generation.
Note that if $w_t$ is an commonsense concept that is out-of-vocabulary, then $p\left(w_{t}\right)$ is zero; if $w_t$ does not exist in the \textit{relevant} commonsense sub-graph, then $y \ast \boldsymbol{\alpha}_{t}^{R}$ is zero.
$x$ and $y$ denote two soft switches to balance between generating
the current word from the baseline distribution $p\left(w_{t}\right)$, or from the attention distribution $\boldsymbol{\alpha}_{t}^{E}$ or $\boldsymbol{\alpha}_{t}^{R}$:
\begin{equation}
\begin{aligned}
x = \sigma(\boldsymbol{W}_x [\boldsymbol{m}_{t}^{0};\boldsymbol{m}_{t}]) \\
y = \sigma(\boldsymbol{W}_y [\boldsymbol{m}_{t}^{0};\boldsymbol{m}_{t}])
\end{aligned}
\end{equation}
where $\boldsymbol{W}_x$ and $\boldsymbol{W}_y$ are learnable matrices;
$\boldsymbol{m}_{t}^{0}$ is the input vector of the first layer of transformer module in Eq. \ref{eq:trans_layer}. 

%
%

%% file: supplement/fig_framework.tex
\begin{figure*}
	\centering{\includegraphics[width=2.0\columnwidth]{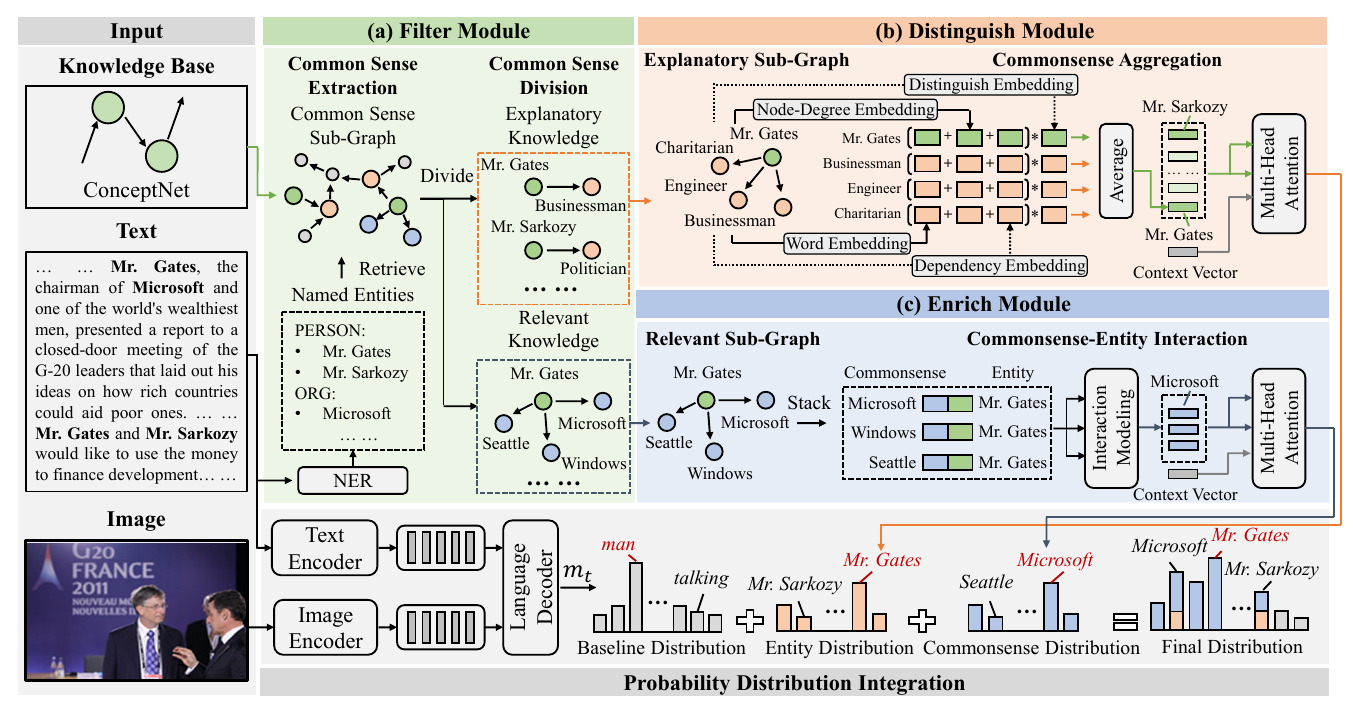}}
	\vspace{-0.1cm}
	\caption{
		Overview of the proposed method.
		We design three communicative modules to exploit commonsense knowledge for the named entity understanding:
		(a) \textbf{Filter Module} queries each named entity of the news article in ConceptNet and obtains the commonsense knowledge, which is divided into \textit{explanatory knowledge} and \textit{relevant knowledge} for the subsequent modules.
		(b) \textbf{Distinguish Module} enhances the entity representations by aggregating the explanatory knowledge from \textit{node-degree}, \textit{dependency}, \textit{distinguish} three aspects, which benefits distinguishing semantically similar named entities (e.g., ``Mr. Gates'' and ``Mr. Sarkozy'').
		(c) \textbf{Enrich Module} models the commonsense-entity interaction based on the relevant knowledge to attach the reliable commonsense concepts for enriching the entity description (e.g., ``Bill Gates, chairman of Microsoft'').
		Finally, the probability distributions from both modules are integrated to generate the news caption.
	}
	\label{fig:framework}
\end{figure*}

%% file: supplement/tab_relations.tex
\begin{table*}[!t]
	\centering
	\caption{Relations in ConceptNet are mainly categorized into two high-level types.}
	\vspace{-0.1cm}
	\scalebox{1.0}{
		\begin{tabular}{c|cccccc}
			\toprule
			\textbf{Category} & \multicolumn{6}{c}{\textbf{Relations}} \\
			\midrule
			\multirow{2}{*}{\textbf{Explanatory Relations}}
			& IsA & FormOf & MannerOf & Synonym & DefinedAs & DerivedFrom \\
			& \multicolumn{2}{c}{EtymologicallyRelatedTo} & \multicolumn{2}{c}{EtymologicallyDerivedFrom} \\
			\midrule
			\multirow{4}{*}{\textbf{Relevant Relations}}
			& RelatedTo & SimilarTo & AtLocation & LocatedNear & CapableOf & Causes \\
			& CausesDesire & CreatedBy & Desires & HasA & HasContext & HasSubevent \\
			& HasProperty & MadeOf & MotivatedByGoal & PartOf & ReceivesAction & SymbolOf \\
			& UsedFor & HasFirstSubevent & HasLastSubevent & HasPrerequisite & & \\
			\midrule
			\textbf{Others}	& Antonym & DistinctFrom & ExternalURL & ObstructedBy \\
			\bottomrule
	\end{tabular}}
	\label{tab:relations}
\end{table*}

%% file: supplement/fig_filter_module.tex
\begin{figure}
	\centering{\includegraphics[width=0.8\columnwidth]{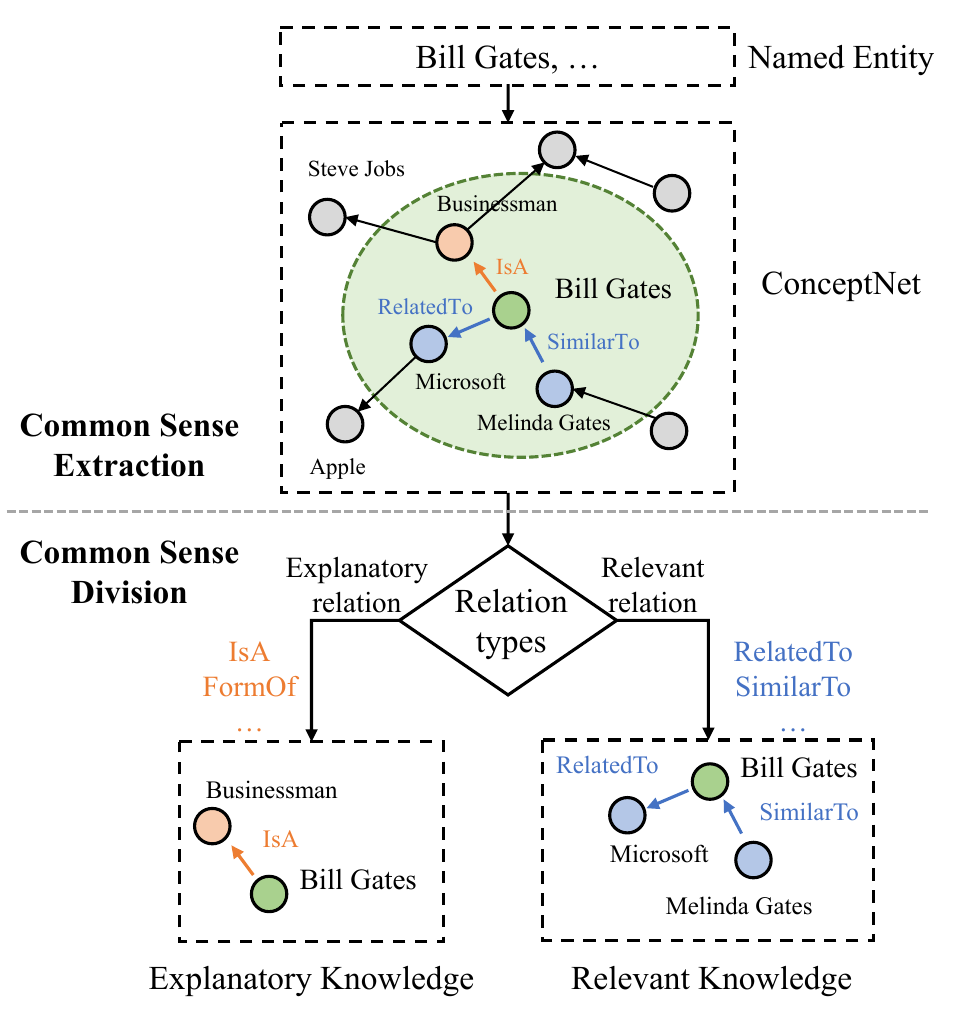}}
	\vspace{-0.1cm}
	\caption{The filter module first queries each named entity in ConceptNet to extract the commonsense knowledge, which is then divided into two sub-graphs by the high-level relation types.}
	\label{fig:filter}
	\vspace{-0.3cm}
\end{figure}

%% file: supplement/fig_distinguish_module.tex
\begin{figure}
	\centering{\includegraphics[width=1.0\columnwidth]{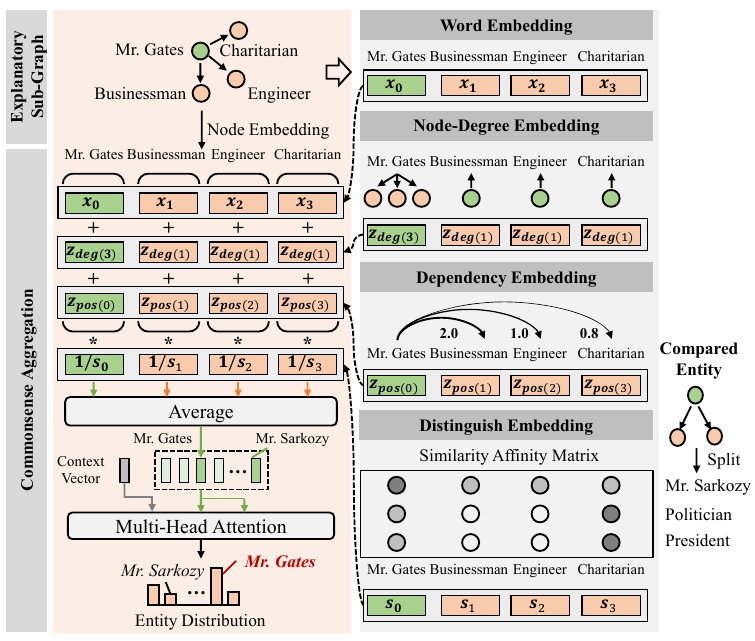}}
	\caption{
	Procedure of Distinguish Module. 
	It aggregates explanatory sub-graph from \emph{node-degree}, \emph{dependency}, and \emph{distinguish} aspects to enhance the entity representation.
	Green denotes the named entity and orange refers to the named entity's explanatory concepts (Best viewed in color).
	}
	\label{fig:distinguish}
	\vspace{-0.3cm}
\end{figure}

%% file: sec_experiment.tex
\section{Experiment} \label{sec:experiment}
\subsection{Datasets and Metrics}
\noindent\textbf{Datasets.} We evaluate the proposed method on two datasets: GoodNews \cite{BitenGRK19} and NYTimes \cite{TranMX20}.
The news articles, images, and captions of both datasets are collected from New York Times, while each image has been annotated by journalists with one caption.
The GoodNews dataset consists of 297,049 articles and 489,168 images in total.
For fair comparison, we adopt the popular split provided by \cite{BitenGRK19}, with a training set of 445,259 samples, a validation set of 19,448 samples and a test set consisting of 24,461 samples.
The NYTimes dataset contains 444,914 articles and 792,971 images in total.
Following the evaluation protocol in \cite{TranMX20}, we use 21,977 samples for validation, 7,777 samples for testing, and 763,217 samples for training.

\noindent\textbf{Metrics.}
We report the performance with BLEU@N (B@N) \cite{PapineniRWZ02}, METEOR (M) \cite{BanerjeeL05}, ROUGE-L (R) \cite{Flick04}, and CIDEr-D (C) \cite{VedantamZP15}.
Specifically, CIDEr is more suited to evaluate the news captioning task than other metrics, since it devalues the stop words and computes more weights on the informative words, including named entities, by a TF-IDF mechanism.
Besides, we report Precision (P), Recall (R), and F1 scores for named entities and rare proper nouns, which are the nouns only appearing in test set but not in training set.
We use SpaCy \cite{honnibal2017spacy} to recognize the named entities in both generated and ground-truth news captions.


\input{./supplement/tab_comparison}

\subsection{Setting}

\noindent\textbf{Pre-processing.}
For the image representation, we extract $N^p$ = 49 image patches and the dimension of each patch feature is set as 2,048. We pick out the Top-4 people's faces and the Top-64 objects with the high confidence scores from the corresponding detectors, while the dimension of face and object feature is set as 512 and 2,048, respectively.
For the article representation, we use the RoBERTa model \cite{abs-1907-11692}, a pretrained language representor with 24 bidirectional transformer layers, where the hidden size and the number of heads in each self-attention block are set to 1,024 and 16, respectively.
Moreover, RoBERTa uses byte-pair-encoding \cite{SennrichHB16a} to build an open vocabulary containing 50,265 subword units, which can handle any unseen word in the named entities as a sequence of subword units.
To reduce the computational cost, we set the maximum length of article sequences as 512 subwords.
The dimension of each subword embedding is set as 1,024.


\noindent\textbf{Commonsense knowledge.}
In the filter module, we select the top 40 named entities in order of appearance in a news article.
To reduce the computational cost, we restrict the entities' \textit{explanatory} and \textit{relevant} commonsense sub-graphs to reserving the top 5 concepts with the high weight values. Specially, these weight values are provided by ConceptNet, which measure the co-occurrence frequency between commonsense concepts and named entities.
In the distinguish and enrich modules,
the dimension of commonsense concept feature, named entity feature, node-degree
embedding, dependency embedding, and the hidden size of multi-head attention are all set to 1,024.
We employ 16 attention heads with the dropout rate 0.1.

\noindent\textbf{Implementation.}
We freeze the parameters of the pretrained models including ResNet, MTCNN, FaceNet, and RoBERTa, and then co-train the entire model.
We use the Adam optimizer \cite{KingmaB15} to compute the gradient with the learning rate of $10^{-4}$.
Following the strategy in \cite{VaswaniSPUJGKP17}, we first warm up the learning rate to $10^{-4}$ and then decrease it linearly.
$L_2$ regularization is applied with a weight decay of $10^{-5}$ to avoid overfitting.
We set the batch size as 16 and train our full model for 22 epochs on GoodNews and 13 epochs on NYTimes.
Our work is implemented in PyTorch \cite{paszke17} using the AllenNLP framework \cite{abs-1803-07640}.
We adopt the mixed-precision training strategy supported by apex\footnote{https://github.com/NVIDIA/apex} to accelerate the training process and reduce the memory footprint.
{Our full model requires approximately 120 hours (5 days) of training time on the GoodNews dataset using a single RTX3090 GPU. On the NYTimes dataset, it takes around 160 hours (7 days) for training.} 

\input{./supplement/tab_ablation}

\input{sec_comparasion}

\input{./supplement/tab_integration_strategy}

\input{sec_ablation}

\input{sec_qualitative}


%% file: supplement/tab_comparison.tex
\begin{table*}[!t]
	\centering
	\caption{
		Performance on GoodNews and NYTimes datasets.
		All values are reported as percentage.}
	\vspace{-0.1cm}
	\begin{tabular}{ll|ccccc|cccccccc}
		\toprule
		& \multirow{2}{*} {Model}
		& \multirow{2}{*} {B@1}
		& \multirow{2}{*} {B@4}
		& \multirow{2}{*} {R}
		& \multirow{2}{*} {M}
		& \multirow{2}{*} {C}
		& \multicolumn{3}{c}{Named Entities}
		& \multicolumn{3}{c}{Rare Proper Nouns} \\
		&&&&&&& P & R & F1 & P & R & F1 \\
		\midrule
		\multirow{12}{*}{\rotatebox[origin=c]{90}{GoodNews}}
		& SAT + CtxIns \cite{XuBKCCSZB15}					& -		& 0.38 & 10.50 & 3.56  & 12.09 & -	   & -	   & 6.09	& -	  	& -	  	& -	\\
		& TopDown + CtxIns \cite{00010BT0GZ18}				& -		& 0.70 & 11.09 & 3.81  & 13.38 & -	   & -	   & 6.87	& -	  	& -	  	& -	\\
		& Ramisa et al. + CtxIns \cite{RamisaYMM18}			& -		& 0.89 & 12.09 & 4.45  & 15.35 & -	   & -	   & 7.54	& -	  	& -	  	& -	\\
		& Biten et al. + CtxIns \cite{BitenGRK19}			& -		& 0.69 & 11.70 & 4.14  & 13.77 & -	   & -	   & 7.19	& -	  	& -	  	& -	\\
		& Biten et al. + AttIns \cite{BitenGRK19}			& -		& 0.76 & 11.44 & 4.02  & 13.69 & -	   & -	   & 7.97	& -	  	& -	  	& -	\\
		& ICECAP \cite{HuCJ20}								& - 	& 1.96 & 15.70 & 6.01  & 26.08 & -	   & -	   & 12.03	& -	  	& -	  	& -	\\
		& Alasdair et al. (Weighted Roberta) \cite{TranMX20}	& 21.43 & 6.00 & 21.17 & 10.14 & 53.07 & 21.76 & 18.53 & 20.02 & 16.19 & 25.97 & 19.95 \\
		& Alasdair et al. (Full Model) \cite{TranMX20}			& 21.42 & 6.05 & 21.43 & 10.26 & 53.81 & 22.24 & 18.68 & 20.31 & 15.63 & 26.25 & 19.59 \\
		& Zhao et al. \cite{abs-2107-11970}						& - 	& 6.14 & 21.46 & 6.32  & 54.02 & - & - & - & - & - & - \\
		& Liu et al. \cite{LiuWWO21}						& - 	& 6.10 & 21.60 & 8.30  & 55.40 & 22.90 & 19.30 & 20.95 & - & - & - \\
		\cmidrule{2-13}
		& \textbf{Ours (Weighted Roberta)}
		& \textbf{22.95} & \textbf{6.26} & \textbf{21.61} & \textbf{10.38} & \textbf{55.40} & \textbf{22.91} & \textbf{19.33} & \textbf{20.97}
		& \textbf{18.92} & \textbf{31.41} & \textbf{23.62} \\
		& \textbf{Ours (Full Model)}
		& \textbf{23.18} & \textbf{6.60} & \textbf{21.77} & \textbf{10.59} & \textbf{56.73} & \textbf{22.92} & \textbf{19.61} & \textbf{21.14}
		& \textbf{19.23} & \textbf{31.34} & \textbf{23.84} \\
		
		\midrule
		\multirow{5}{*}{\rotatebox[origin=c]{90}{NYTimes}}
		& Alasdair et al. (Weighted Roberta) \cite{TranMX20}& 20.92 & 5.75 & 19.89 &  9.56 & 45.12 & 21.06 & 19.61 & 20.31 & 29.59 & 22.77 & 25.74 \\
		& Alasdair et al. (Full Model) \cite{TranMX20}		& 21.63 & 6.30 & 21.65 & 10.34 & 54.43 & 24.57 & 22.15 & 23.30 & 34.19 & 27.00 & 30.17 \\
		& Zhao et al.  \cite{abs-2107-11970}				& - 	& 6.32 & 21.62 &  6.25 & 54.47 & -     & -     & 23.44 & -     & -     & -     \\
		& Liu et al. \cite{LiuWWO21}						& - 	& 6.40 & 21.90 & 8.10  & 56.10 & 24.80 & 22.30 & 23.48 & - & - & - \\
		\cmidrule{2-13}
		& \textbf{Ours (Full Model)}
		& \textbf{22.61} & \textbf{6.94}  & \textbf{22.04} & \textbf{10.74} & \textbf{57.27} & \textbf{25.05} & \textbf{23.08} & \textbf{24.02} & \textbf{37.86} &\textbf{31.40}  & \textbf{34.33} \\
		\bottomrule
	\end{tabular}
	\label{tab:comparison}
\end{table*}

%% file: supplement/tab_ablation.tex
\begin{table*}[!t]
	\centering
	\footnotesize
	\caption{
		Performance of variants for the proposed method on GoodNews and NYTimes datasets.}
	\vspace{-0.1cm}
	\renewcommand\arraystretch{1.1}
	\scalebox{0.98}{
		\begin{tabular}{ll|ccccc|cccccc}
			\toprule
			& \multirow{2}{*} {Model}
			& \multirow{2}{*} {B@1}
			& \multirow{2}{*} {B@4}
			& \multirow{2}{*} {R}
			& \multirow{2}{*} {M}
			& \multirow{2}{*} {C}
			& \multicolumn{3}{c}{Named Entities}
			& \multicolumn{3}{c}{Rare Proper Nouns} \\
			&&&&&&& P & R & F1 & P & R & F1 \\
			\midrule
			\multirow{4}{*}{\rotatebox[origin=c]{90}{GoodNews}}
			& Non-Commonsense    				
			& 21.42 & 6.05 & 21.43 & 10.26 & 53.81 & 22.24 & 18.68 & 20.31 & 15.63 & 26.25 & 19.59 \\
			& Non-Distinguish    				
			& 22.91 &  6.38 & 21.59 & 10.47 & 56.01	& 22.28	& 19.16	& 20.60 & 19.10	& 30.04 & 23.26 \\
			& Non-Enrich   				
			& 22.70 &  6.34 & 21.68 & 10.43 & 56.42 & 22.53 & 19.25 & 20.76 & 18.82 & 29.74 & 23.14 \\
			& \textbf{Ours (Full Model)}   				
			& \textbf{23.18} &  \textbf{6.60} & \textbf{21.77} & \textbf{10.59} & \textbf{56.73} & \textbf{22.92} & \textbf{19.61} & \textbf{21.14}
			& \textbf{19.23} & \textbf{31.34} & \textbf{23.84} \\
			\midrule
			\multirow{4}{*}{\rotatebox[origin=c]{90}{NYTimes}}
			& Non-Commonsense    				
			& 21.63 & 6.30 & 21.65 & 10.34 & 54.43 & 24.57 & 22.15	& 23.30 & 34.19 & 27.00 & 30.17 \\
			& Non-Distinguish   				
			& 22.34 &  6.79 & 21.91 & 10.60 & 57.09	& 24.96	& 22.86	& 23.86 & 37.29	& 30.09 & 33.31 \\
			& Non-Enrich   				
			& 21.58 &  6.49 & 21.72 & 10.37 & 57.18 & 25.02 & 22.91 & 23.92 & 37.16 & 29.87 & 33.12 \\
			& \textbf{Ours (Full Model)}   				
			& \textbf{22.62} & \textbf{6.94} & \textbf{22.04} & \textbf{10.74} & \textbf{57.27} & \textbf{25.05} & \textbf{23.08} & \textbf{24.02} & \textbf{37.86} &\textbf{31.40}  & \textbf{34.33} \\
			
			\bottomrule
		\end{tabular}}
	\label{tab:ablation}
\end{table*}

%% file: sec_comparasion.tex
\subsection{Comparison with the State-of-The-Art}
We compare the proposed method to two types of news captioning models.
1) \textbf{Template-based models.}
It first produces template captions with named entities' placeholders, and then selects the best candidate to fill each placeholder for news captioning generation \cite{XuBKCCSZB15,00010BT0GZ18,RamisaYMM18,BitenGRK19,HuCJ20}.
For examples,
Biten \textit{et al}. \cite{BitenGRK19} proposes to choose named entities with attention values to fill the template captions.
Hu \textit{et al}. \cite{HuCJ20} optimizes the template caption generation and the named entity selection in an end-to-end manner.
2) \textbf{Joint learning models.}
It directly generates the named entities and linguistically rich sentences without the selection process and template captions \cite{TranMX20,abs-2107-11970,LiuWWO21}.
For examples,
Alasdair \textit{et al}. \cite{TranMX20} associate entities with faces and objects of an image, which uses byte-pair-encoding to generate sentences.
Zhao \textit{et al}. \cite{abs-2107-11970} build a multi-modal knowledge graph to connect the textual entities in news articles and visual objects of images.
Liu \textit{et al}. \cite{LiuWWO21} equip the transformer architecture with multi-modal fusion and attention mechanisms to leverage images, articles, and entities.

{\color{magenta}}
As shown in Tab. \ref{tab:comparison}, the proposed method outperforms both template-based models and joint learning models, which can achieve competitive performance on both GoodNews and NYTimes datasets.
We have four key observations:
\begin{itemize}
\item Our method consistently exhibits better performance than the aforementioned state-of-the-art methods in terms of all metrics. 
    Especially, F1 score of rare proper nouns can achieve 23.84\% on GoodNews and 34.33\% on NYTimes, which 
    makes the absolute improvement over the best competitor Alasdair \textit{et al}. (Full Model) \cite{TranMX20} by 4.25\% and 4.16\%, respectively. It validates the superiority of using the commonsense knowledge to distinguish and describe named entities for news captioning.

\item Our method can significantly outperform the template-based models, which directly copy the named entities from articles and take no account of entity understanding. Specifically, the proposed enrich module can analyze the commonsense knowledge and extract relevant semantics to completely describe a named entity. As shown in Tab. \ref{tab:comparison}, our method achieves a recall of 31.41\% on rare proper nouns, which illustrates that commonsense outside of articles can be effectively introduced to enrich the language corpus for news image description.

\item We compare to the jointly learning methods, which mainly depend on the context cues, such as person's face, visual object, and textual word, to produce named entities. Comparatively, our distinguish module uses commonsense knowledge beyond the context to explore the similar named entities, which can benefit generating entities more accurately. Tab. \ref{tab:comparison} shows that the proposed method can significantly improve the performance over the strong baseline Liu \textit{et al}. \cite{LiuWWO21} on both datasets. It confirms the superiority of distinguishing similar entities for news caption generation.

\item For the fair comparison with other methods, we follow Alasdair \textit{et al}. \cite{TranMX20} and remove the features of objects and faces in an image, which is denoted as Ours (Weighted Roberta). As shown in Tab. \ref{tab:comparison}, even without object and face representations, Ours (Weighted Roberta) still outperforms all other comparison methods including Alasdair \textit{et al}. (Full Model) \cite{TranMX20} on GoodNews. It further verifies the advantage of integrating commonsense knowledge.
	
\end{itemize}

%% file: supplement/tab_integration_strategy.tex
\begin{table}[!t]
	\setlength{\tabcolsep}{3pt}
	\footnotesize
	\centering
	\caption{Comparison of variants to measure the effectiveness of commonsense division on NYTimes.}
	\begin{tabular}{ll|ccccccc}
		\toprule
		& \multirow{2}{*} {Model}
		& \multirow{2}{*} {C}
		& \multicolumn{3}{c}{Named Entities}
		& \multicolumn{3}{c}{Rare Proper Nouns} \\
		&&& P & R & F1 & P & R & F1 \\
		\midrule
		& Non-Division					& 56.05 & 24.53 & 22.55 & 23.50 & 37.48 & 31.00 & 33.93  \\
		& Division$\star$               & 56.62 & 24.90 & 22.86 & 23.84 & 37.63 & 31.15 & 34.08  \\
		& \textbf{Ours}  										
		& \textbf{57.27} & \textbf{25.05} & \textbf{23.08} & \textbf{24.02} & \textbf{37.86} &\textbf{31.40}  & \textbf{34.33} \\
		
		\bottomrule
	\end{tabular}
	\label{tab:ablation-2}
	\vspace{-0.1cm}
\end{table}

%% file: sec_ablation.tex
\subsection{Ablative Analysis}

\input{./supplement/fig_line_chart}

We conduct a number of ablations to analyze the proposed method by answering the following questions.
\textbf{Q1}: Is the idea of using commonsense knowledge helpful for the task of news captioning?
\textbf{Q2}: Is it necessary to divide commonsense into the \textit{explanatory} and \textit{relevant} categories in the filter module?
\textbf{Q3}: What are the effects of the \textit{distinguish} and \textit{enrich} modules for named entity understanding?
Results are reported in Table \ref{tab:ablation} and discussed as follows.

\noindent\textbf{Effectiveness of Using Commonsense Knowledge (Q1):}
We exploit how to use commonsense to boost the news captioning task, where a series of commonsense-aware modules, including the \textit{filter}, \textit{distinguish}, and \textit{enrich} modules, are designed to retrieve knowledge database and integrate commonsense for the named entity understanding.
We validate the effectiveness from the following two aspects.

\noindent\textbf{1) Non-Commonsense.}
We design the variant \textit{Non-Commonsense} by removing all of the proposed commonsense-aware modules.
Specifically, this variant abandons the entity distribution $\boldsymbol{\alpha}_{t}^{E}$ and the commonsense distribution $\boldsymbol{\alpha}_{t}^{R}$ in Eq. \ref{eq:fwd} while generating the final word distribution depending on the context representation $\boldsymbol{m}_{t}$ as follows:
\begin{equation}
p^*\left(w_t\right)= \operatorname{AdaptiveSoftmax}\left(\boldsymbol{W}_p \boldsymbol{m}_{t}\right)
\end{equation}
where Non-Commonsense falls into the baseline model similar to Alasdair et al. \cite{TranMX20}.
As shown in Tab. \ref{tab:ablation},
after integrating commonsense knowledge, Ours (Full Model) can consistently outperform Non-Commonsense in terms of all metrics.
On both datasets, the absolute improvements are about 3\% under the CIDEr score.
It indicates that the commonsense can provide facts beyond the given article and image to help the model understand the meaning of named entities.
The results are consistent with our intuitions mentioned in the introduction.

\noindent\textbf{2) Investigation on the Number of Commonsense Concepts.}
In the filter module, we extract the top-$K$ named entities from the news article, which are used as the queries to retrieve commonsense concepts (i.e., sub-graphs) from ConceptNet.
Intuitively, the more named entities used for commonsense retrieval, the richer concept knowledge available to the model.
To investigate the impact of the number of commonsense concepts on the performance, we control the number of queried entities $K$ from 0 to 50 in Fig. \ref{fig:line_chart}.
Particularly, in this paper, we focus on how to understand named entities via commonsense. Therefore, we use the  Precision, Recall, and F1 scores of named entities in the generated captions as metrics.
Besides, to clearly compare the performance with the different number of queried entities, we normalize the evaluation  scores as follows:
\begin{equation}
\tilde{v}_K=\frac{v_K-\min_K\left\{v_K\right\}}{\min_K\left\{v_K\right\}}
\end{equation}
where $v_K$ is the Precision, Recall, or F1 score of querying $K$ named entities;
$\tilde{v}_K$ denotes the normalized performance value.

As shown in Fig. \ref{fig:line_chart}, all of the curves gradually increase when $K$ ranges from 0 to 40.
It is because the abundant commonsense concepts can help distinguish the similar entities and enrich the entity description.
However, when $K$ ranges from 40 to 50, all of the curves will drop since the overmuch commonsense may bring irrelevant noise and have a negative impact on the performance.
In our experiments, the number of queried named entities $K$ is empirically set as 40.

\noindent\textbf{Effectiveness of Commonsense Division (Q2):}
In order to seek the informative concepts to distinguish and describe named entities, we divide the commonsense sub-graph into the \textit{explanatory} and \textit{relevant} ones, respectively.
The former is fed into the distinguish module while the latter is used by the enrich module.
To validate the effectiveness, we adopt the following methods of 1) removing the process of division and 2) comparing with another division strategy.

\noindent\textbf{1) Non-Division.}
We compare the proposed method with its variant \textit{Non-Division} that does not divide the retrieved commonsense knowledge and directly feeds them into both the distinguish and enrich modules.
From Tab. \ref{tab:ablation-2}, we can observe that our model consistently improves the performance and outperforms Non-Division under the named entities and the rare proper nouns.
It is because the process of commonsense division can provide the accurate and subtle semantic space to restrict distinguish or enrich module for the named entity understanding.
Meanwhile, Non-Division is only indiscriminately accepting all of retrieved commonsense, which may introduce noisy semantic for each module. 

\input{./supplement/fig_visualization}

\noindent\textbf{2) Investigation on Division Strategy.}
As shown in Tab. \ref{tab:relations}, we categorize the relations of ConceptNet into the \textit{explanatory} and \textit{relevant} ones, both of which are used to divide the retrieved commonsense into two sub-graphs for the corresponding modules.
{To more clearly capture the commonsense semantics,
we design the variant \textit{Division$\star$} that only selects the most semantically relevant relation ``IsA'' and ``RelatedTo'' to build sub-graphs.
Specifically, we construct the \textit{explanatory} sub-graph using the commonsense concepts linked by the ``IsA'' relation in Microsoft Concept Graph dataset \cite{WangWWX15,WuLWZ12}, which can provide richer triplets of ``IsA'' relations compared to ConceptNet. For the \textit{relevant} sub-graph, we continue to use the commonsense concepts connected through the ``RelatedTo'' relation in ConceptNet.}

Table \ref{tab:ablation-2} shows that Division$\star$ can achieve better performances than Non-Division, which proves the efficiency of commonsense division once again.
Besides, it also illustrates that Microsoft Concept Graph can provide the \textit{explanatory} factual knowledge to help distinguish similar entities.
Moreover, our model performs better than Division$\star$.
The main difference is that we only use one data source ConceptNet while Division$\star$ uses two sources including ConceptNet and Microsoft Concept Graph.
The domain gap between sources could have a negative impact on the Division$\star$'s performance.
It demonstrates that how to integrate the different commonsense sources for named entity understanding is a challenging problem in future works.

\noindent\textbf{Effectiveness of the Distinguish and Enrich Modules (Q3):}
The \textit{distinguish} and \textit{enrich} modules are two parallel processes.
The former depends on commonsense to distinguish semantically similar named entities.
The latter captures relevant concepts from commonsense which are beyond the given article, to describe a named entity completely.
The probability distributions from both modules are integrated to generate news captions.
For quantifying their importance, we conduct two ablations:
1) \textbf{Non-Distinguish:}
We remove the distinguish module with its entity distribution $\boldsymbol{\alpha}_{t}^{E}$ and Eq. \ref{eq:fwd} is modified as:
\begin{equation}
p^*\left(w_t\right)= y \ast \boldsymbol{\alpha}_{t}^{R} + (1-y) \ast p\left(w_t\right)
\end{equation}
2) \textbf{Non-Enrich:}
We remove the enrich module with its commonsense distribution $\boldsymbol{\alpha}_{t}^{R}$ and Eq. \ref{eq:fwd} is modified as:
\begin{equation}
p^*\left(w_t\right)= x \ast \boldsymbol{\alpha}_{t}^{E} + (1-x) \ast p\left(w_t\right)
\end{equation}

From results in Table \ref{tab:ablation}, we obtain four key points:
\begin{itemize}
\item Both of \textit{Non-Distinguish} and \textit{Non-Enrich} can help to improve the performance over \textit{Non-Commonsense}, which shows the effectiveness of using commonsense to distinguish similar named entities and enrich the descriptions of entities, respectively. Specially, under the F1 score of rare proper nouns, the improvements are over 18.0\%.

\item \textit{Non-Enrich} performs better than \textit{Non-Distinguish} under all scores of named entities. It suggests that the proposed distinguish module can allow the captioner to generate named entities accurately by aggregating \textit{explanatory} commonsense for similar entities.

\item \textit{Non-Distinguish} outperforms \textit{Non-Enrich} under all scores of rare proper nouns, which evaluates the nouns that only appear in test set but not in training set. It indicates that the proposed enrich module can capture the reliable concepts from \textit{relevant} commonsense outside of articles to describe named entities.

\item Our method can consistently outperform \textit{Non-Distinguish} and \textit{Non-Enrich}. It suggests that both variants are complementary to each other in our full model, while the abilities to distinguish and describe named entities are compatible and mutually reinforcing for news captioning.
\end{itemize}

%% file: supplement/fig_line_chart.tex
\begin{figure*}[htbp]
	\centering
	\includegraphics[width=2.0\columnwidth]{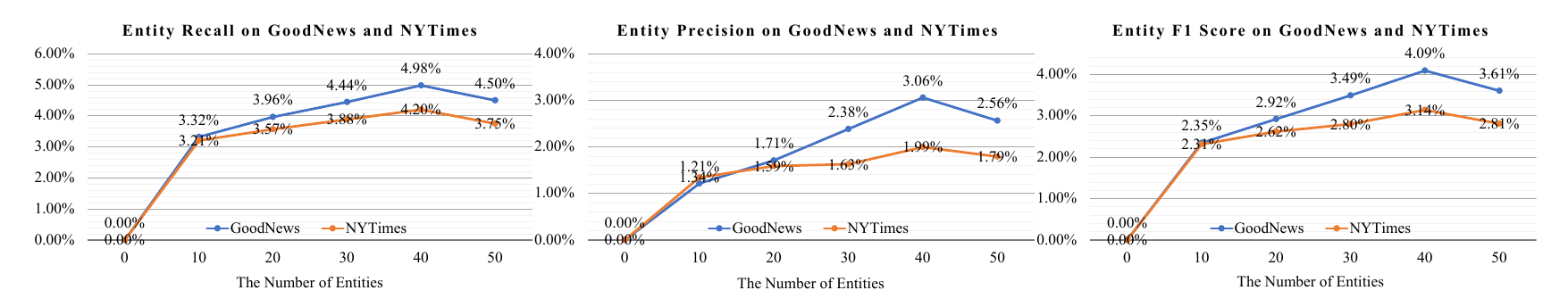}
	\caption{Performance of the proposed method on GoodNews and NYTimes datasets when using different numbers of named entities to retrieve commonsense knowledge.}
	\label{fig:line_chart}
	\vspace{-0.2cm}
\end{figure*}

%% file: supplement/fig_visualization.tex
\begin{figure*}
	\centering{\includegraphics[width=2.0\columnwidth]{./figs/visualization-3.pdf}}
	\caption{
	{Qualitative results of our method on GoodNews dataset.
	For each example, we present several detected entities in news articles and the corresponding \textit{explanatory} and \textit{relevant} commonsense knowledge.
The key named entities and commonsense concepts are underlined to examine their role in generating the named entities. The inaccurate generated named entities are marked in red, while the accurate named entities are indicated in green.}
	\textbf{Ground-Truth} denotes the news captions written by journalists.
	\textbf{Baseline} presents the generated sentences from the captioning model released by Alasdair et al. \cite{TranMX20}.
	Compared with the baseline model, our method can integrate the external commonsense knowledge to explicitly distinguish and describe named entities (Best viewed in color).
	}
	\label{fig:visulization}
	\vspace{-0.2cm}
\end{figure*}

%% file: sec_qualitative.tex
\subsection{Qualitative Analysis}\label{qualitative_analysis}
Fig. \ref{fig:visulization} shows four examples with news image captions generated by our method and the baseline model, Alasdair et al. \cite{TranMX20}.
To provide insight in how the commonsense influences the named entity generation, we present several detected entities in news articles and the corresponding \textit{explanatory} and \textit{relevant} commonsense knowledge. 
Meanwhile, we underline the key named entities and commonsense concepts to examine their role in generating named entities. The inaccurate generated named entities are marked in red, while the accurate named entities are indicated in green. 
{The qualitative results demonstrate that the introduced commonsense knowledge significantly enhances the interpretability of our proposed method. The generated news captions effectively describe the named entities of the given images and articles in a reasonable manner.} 
We have the following three observations. 

\begin{itemize}
\item {Depending on the \textit{explanatory} commonsense knowledge, our method can distinguish similar named entities appearing in news articles and decide which entity should be used to describe the image content. For example, in case (a), the baseline model cannot effectively distinguish ``Tony Blair'' and ``Wendi Deng'', and generates the wrong description ``Mr. Murdoch with his wife, Tony Blair''. In contrast, our model can capture the commonsense ``Wendi Deng is Rupert Murdoch's wife'' and increase the possibility of ``Wendi Deng'' in the probability distribution for the more accurate caption generation.}

\item The proposed method can integrate the \textit{relevant} commonsense knowledge to completely describe the named entities. In case (c), our model can associate ``John Kerry'' with the fact ``John Kerry is an office holder who is related to United States Department of State'' and produce the rich description ``Secretary of State John Kerry''. Comparatively, the baseline model fails to generate the detail introduction about ``John Kerry'' due to the lack of commonsense knowledge.

\item We notice a failure case in Fig. \ref{fig:visulization}(d), where our model is unable to output the detail information ``a cancer patient'' for ``Elena''. The reason is that since ``Elena'' is not famous enough, our filter module cannot retrieve her sufficient commonsense from ConceptNet. It may be helpful to integrate the multi-source knowledge such as news, wikis, and web text to enrich the commonsense corpus in future works.

\end{itemize}

%% file: sec_conclusion.tex
\section{Conclusion}

{
In this paper, we propose the novel news captioning method to use the common sense knowledge for named entity understanding. Different from existing works that overlook the importance of common sense knowledge, we define the main challenge of news captioning as leveraging common sense beyond the given news content to understand named entities. Particularly, three communicative modules are designed to divide and integrate the common sense knowledge, which aims to distinguish semantically similar named entities and describe a named entity completely. Comparative results against state-of-the-arts on two popular datasets show the superiority of our method. Ablative experiments and qualitative analysis validate the effectiveness. In the future, we will consider 1) how to acquire richer and more relevant common sense knowledge; 2) how to balance the probability distributions of named entities across people, organizations, and places. 
}

%% file: sec_acknowledgement.tex
\section*{Acknowledgments}


This work was supported in part by the National Natural Science Foundation of China (U21B2024, 62002257). 

\vspace{-1ex}